\documentclass[review]{elsarticle}

\usepackage{lineno, hyperref}
\modulolinenumbers[5]

\journal{Journal of Web Semantics}

\usepackage{url}  
\usepackage{graphicx}  
\usepackage{pdfpages}

\usepackage{color}
\usepackage[normalem]{ulem}

\usepackage{amsfonts}       
\usepackage{amsmath, mathrsfs, amsfonts,amssymb}
\usepackage{booktabs}       

\begin{document}

\begin{frontmatter}

\title{Embedding Models for Episodic Knowledge Graphs}

\author[1,2]{Yunpu Ma}
\author[1,2]{Volker Tresp}
\author[3]{Erik A. Daxberger~\footnote{Work done while at Siemens AG.}}

\address[1]{Siemens AG, Corporate Technology, Munich, Germany}
\address[2]{Ludwig Maximilian University of Munich, Munich, Germany}
\address[3]{ETH Zurich}

\begin{abstract}
In recent years a number of large-scale triple-oriented knowledge graphs have been generated and various models have been proposed to perform learning in those graphs. Most knowledge graphs are static and reflect the world in its current state. In reality, of course, the state of the world is changing: a healthy person becomes diagnosed with a disease and a new president is inaugurated. In this paper, we extend models for static knowledge graphs to temporal knowledge graphs. This enables us to store episodic data and to generalize to new facts (inductive learning). We generalize leading learning models for static knowledge graphs (i.e., Tucker, RESCAL, HolE, ComplEx, DistMult) to temporal knowledge graphs. In particular, we introduce a new tensor model, ConT,
with superior generalization performance. The performances of all proposed models are analyzed on two different datasets: the Global Database of Events, Language, and Tone (GDELT) and the database for Integrated Conflict Early Warning System (ICEWS).
We argue that temporal knowledge graph embeddings might be models also for cognitive episodic memory (facts we remember and can recollect) and that a semantic memory (\emph{current} facts we know) can be generated from episodic memory by a marginalization operation. We validate this episodic-to-semantic projection hypothesis with the ICEWS dataset.
\end{abstract}


\end{frontmatter}


\section{Introduction}

In recent years a number of sizable Knowledge Graphs (KGs) have been developed, the largest ones containing  more than 100  billion facts. Well known examples are DBpedia~\cite{auer2007dbpedia},YAGO~\cite{suchanek2007yago}, Freebase~\cite{bollacker2008freebase}, Wikidata~\cite{vrandevcic2014wikidata} and the Google KG~\cite{singhal2012introducing}. Practical issues with completeness, quality and maintenance have been solved to a degree that some of these knowledge graphs   support search, text understanding and question answering in large-scale commercial systems~\cite{singhal2012introducing}. In addition,  statistical embedding  models have been developed that can be used to compress a knowledge graph, to derive implicit facts, to detect errors, and to support the above mentioned applications. A recent survey on KG models can be found in~\cite{nickel2015}.

Most knowledge graphs are static and reflect the world at its current state. In reality, of course, the state of the world is changing: a healthy person becomes diagnosed with a disease and a new president is inaugurated. In this paper, we extend semantic knowledge graph embedding models to episodic/temporal knowledge graphs as an efficient way to store episodic data and to be able to generalize to new facts (inductive learning). In particular, we generalize leading approaches for static knowledge graphs (i.e., constrained Tucker, DistMult, RESCAL, HolE, ComplEx) to temporal knowledge graphs. We test these  models using two temporal KGs. The first one is derived from the Integrated Conflict Early Warning System (ICEWS) data set which describes interactions between nations over several years. The second one is derived from the Global Database of Events, Language and Tone (GDELT) that, for more than 30 years,  monitors news media from all over the world. In the experiments, we analyze the generalization abilities to new facts that might be missing in the temporal KGs and also analyze to what degree a factorized KG can serve as an explicit memory.

We propose that our technical models might be related to the brain's explicit memory systems, i.e., its episodic and its semantic memory. Both are considered long-term memories and store information potentially over the life-time of an individual~\cite{ebbinghaus1885gedachtnis,atkinson1968human,squire1987memory, ebbinghaus1885gedachtnis}. The semantic memory stores general factual knowledge, i.e., information we \textit{know}, independent of the context where this knowledge was acquired and would be related to a static KG.  Episodic memory concerns information we \textit{remember} and includes the spatiotemporal context of events~\cite{tulving1986episodic} and would correspond to a temporal KG.

\begin{figure*}[htp]
   \centering
   \includegraphics[ width=1.0\textwidth]{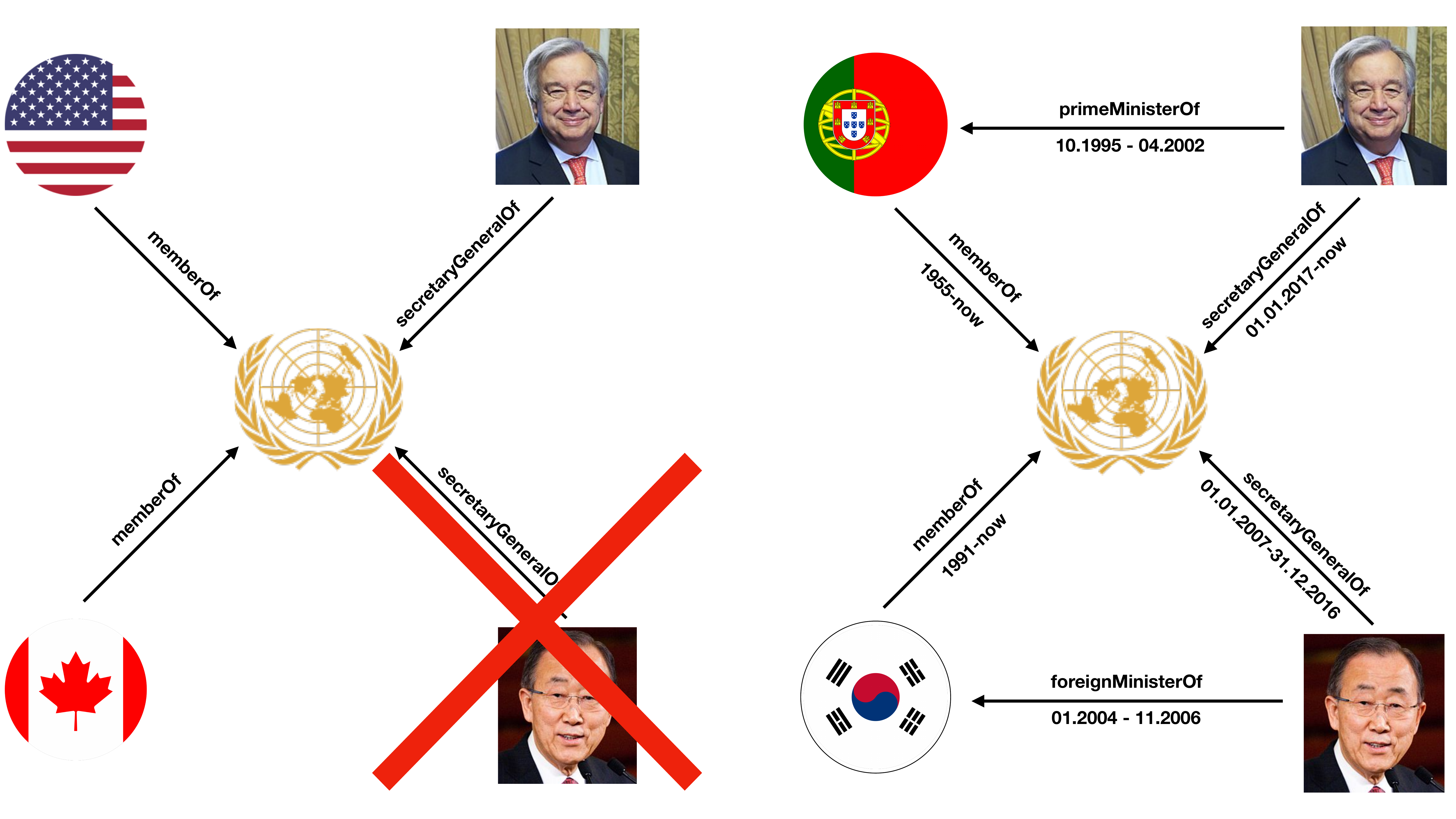}
  \caption{Illustrations of (left) a semantic knowledge graph and (right) an episodic knowledge graph. (Left) Every arrow represents a (subject, predicate, object) triple, with the annotation of the arrow denoting the respective predicate. The triple (Ban Ki-moon, SecretaryOf, UN) is deleted, since the knowledge graph has been updated with the triple (Ant\'onio Guterres, SecretaryOf, UN). (Right) Every arrow represents a (subject, predicate, object, timestamp) quadruple, where the arrow is both annotated with the respective predicate and timestamp. Here the quadruple involving  is not deleted, since the attached timestamp reveals that the relationship is not valid at present.}
  \label{fig:semantic_episodic_kg}
\end{figure*}

\begin{figure*}[htp]
   \centering
   \includegraphics[ width=1.0\textwidth]{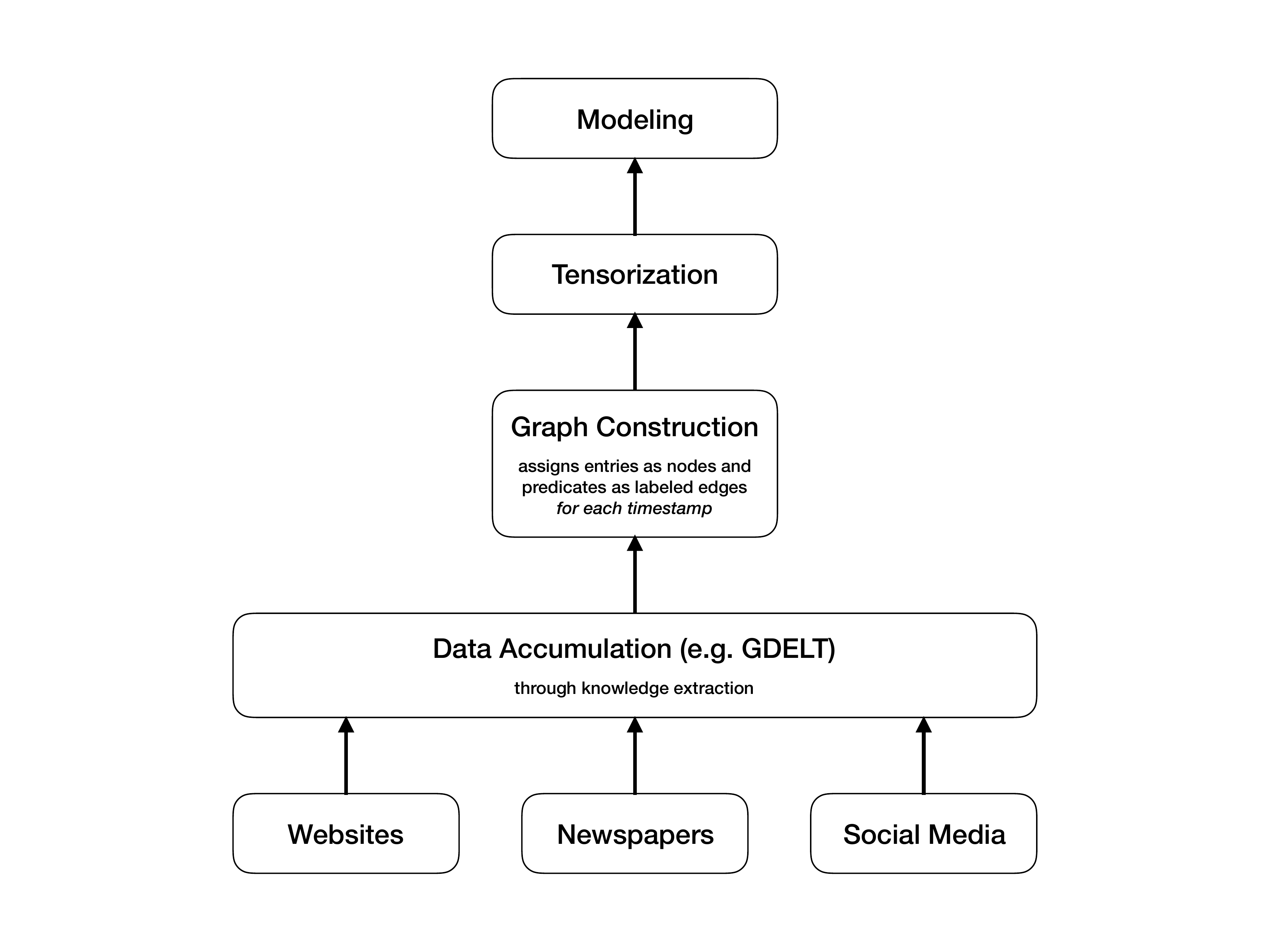} \caption{Illustration of the main idea behind the models presented in this paper. \textbf{Step 1}: Knowledge is extracted from unstructured data, such as websites, newspapers or social media. \textbf{Step 2}: The knowledge graph is constructed, where entities are assigned as nodes, and predicates as labeled edges; note that there is a labeled edge for each timestamp. \textbf{Step 3}: The knowledge graph is represented as a tensor; for semantic KGs, we obtain a 3-way tensor, storing (subject, predicate, object) triples, and for episodic KGs, we obtain a 4-way tensor, storing (subject, predicate, object, timestamp) quadruples. \textbf{Step 4}: The semantic and episodic tensors are decomposed and modeled via compositional or tensor models (see Section~\ref{sec: Models}).}
  \label{fig:idea_of_models}
\end{figure*}

An interesting question is how episodic and semantic memories are related. There is evidence that these main cognitive categories are partially dissociated from one another in the brain, as expressed in their differential sensitivity to brain damage. However,  there is also evidence indicating  that the different memory functions are not mutually independent and support one another~\cite{greenberg2010interdependence}. We propose that semantic memory can be derived from episodic memory by marginalization. Hereby we also consider that many episodes describe starting and endpoints of state changes. For example, an individual might become sick with a disease, which eventually is cured. Similarly, a president's tenure eventually ends. We study our hypothesis on the Integrated Conflict Early Warning System (ICEWS) dataset, which contains many events with start and end dates. Figure~\ref{fig:semantic_episodic_kg} compares semantic and episodic knowledge graphs. Furthermore, Figure~\ref{fig:idea_of_models} illustrates the main ideas of building and modeling semantic and episodic knowledge graphs.

The paper is organized as follows. Section~\ref{sec: Models} introduces knowledge graphs, the mapping of a knowledge graph to an adjacency tensor, and the statistical embedding models for knowledge graphs. We also describe how popular embedding models for KGs can be extended to episodic KGs. Section~\ref{sec:Episodic_exp} shows experimental results on modelling episodic KGs. Finally, we present experiments on the possible relationships between episodic and semantic memory in Section~\ref{sec:Projection_exp}.

\section{Model Descriptions}
\label{sec: Models}

A   static or semantic knowledge graph (KG)  is a triple-oriented knowledge representation.  Here we consider a slight extension to    the subject-predicate-object triple form by adding the value in the form ($e_s, e_p, e_o$; \textit{Value}), where \textit{Value} is a function of $e_s, e_p, e_o$ and, e.g.,  can be a Boolean variable (\textit{True} for \textit{1}, \textit{False} for \textit{0})  or a real number.
Thus \textit{(Jack, likes, Mary; True)} states that Jack (the subject or head entity) likes Mary (the object or tail entity).
Note that $e_s$ and $e_o$ represent the entities for subject index $s$ and object index $o$. To simplify notation we also consider $e_p$ to be a generalized  entity associated with predicate type with index $p$.
For the episodic KGs we introduce $e_t$, which is  a generalized entity for time $t$.

To model a static KG,  we introduce the three-way semantic adjacency tensor $\chi$ where the tensor element $x_{s, p, o}$ is the associated \emph{Value} of the triple $(e_s, e_p, e_o)$.  One can also define a companion tensor $\Theta_{\chi}$ with the same dimensions as $\chi$ and with entries $\theta_{s, p, o}$. Thus, the probabilistic model for the semantic tensor $\chi$ is defined as $  P(x_{s,p,o}|\theta_{s,p,o}) = \sigma(\theta_{s,p,o})$,
where $\sigma(x)=1/(1+\exp(-x))$.
Similarly, the four-way temporal or episodic tensor $\mathcal{E}$ has elements $x_{t, s, p, o}$ which are the associated values of the quadruples $(e_t, e_s, e_p, e_o)$, with $t=1,\dots, T$. Therefore, the probabilistic model for episodic tensor is defined with the corresponding companion tensor $\Theta_{\mathcal{E}}$ as
\begin{equation}
  P(x_{t, s, p, o}|\theta_{t, s, p, o}) = \sigma(\theta_{t, s, p, o})\ .
\end{equation}
We assume that each entity $e$ has a unique latent representation $\mathbf{a}$. In particular, the embedding approach used for modeling semantic and episodic knowledge graphs assumes that $  \theta^{sem}_{s,p,o} = f^{sem}(\mathbf{a}_{e_s}, \mathbf{a}_{e_p}, \mathbf{a}_{e_o})$, and $\theta^{epi}_{t, s,p,o} = f^{epi}(\mathbf{a}_{e_t}, \mathbf{a}_{e_s}, \mathbf{a}_{e_p}, \mathbf{a}_{e_o})$, respectively. Here, the indicator function $f^{sem/epi}(\cdot)$ is a  function to be learned.

Given a labeled dataset $\mathcal{D}=\{(x_i, y_i)\}_{i=1}^{m}$, latent representations and other parameters (denoted as $\mathcal{P}$) are learned by minimizing the regularized logistic loss
\begin{equation}
  \min_{\mathcal{P}}\sum_{i=1}^m \log (1+\exp(-y_i\theta^{sem/epi}_i)) + \lambda ||\mathcal{P}||_2^2.
  \label{logisticloss}
\end{equation}
In general, most KGs only contain positive triples; non-existing triples are normally used as negative examples sampled with local closed- world assumption. Alternatively, we can minimize a margin-based ranking loss over the dataset such as
\begin{equation}
  \min_{\mathcal{P}}\sum_{i\in\mathcal{D}_{+}}\sum_{j\in\mathcal{D}_{-}}\max(0, \gamma + \sigma (\theta^{sem/epi}_j) - \sigma (\theta^{sem/epi}_i)),
  \label{rankingloss}
\end{equation}
where $\gamma$ is the margin parameter, and $\mathcal{D}_{+}$ and $\mathcal{D}_{-}$ denote the set of positive and negative samples, respectively.

\begin{figure*}[htp]
   \centering
  {\begin{tabular}{ccccc}
	  \hspace{-5mm} \includegraphics[page=1, width=0.22\textwidth]{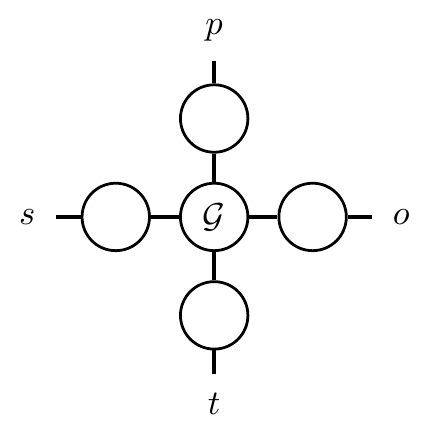}
	& \hspace{-5mm} \includegraphics[page=2, width=0.22\textwidth]{Figure/Tensor_Graphics}
	& \hspace{-5mm} \includegraphics[page=4, width=0.14\textwidth]{Figure/Tensor_Graphics}
	& \includegraphics[page=3, width=0.14\textwidth]{Figure/Tensor_Graphics}
	& \includegraphics[page=5, width=0.18\textwidth]{Figure/Tensor_Graphics} \\
	(a) & (b) & (c) & (d) & (e)
  \end{tabular}}
  \caption{Illustrations of (a) episodic Tucker, (b) episodic ComplEx (where $\bullet$ denotes contraction), (c) RESCAL, (d) ConT and (e) Tree. Each entity in the figure is represented as a circle with two edges, since the representation for an entity $e$ is $\mathbf{a}_{e, i}$. In addition, $\mathcal{G}$ represents the core tensor in Tucker, $\mathcal{G}_p$ represents the matrix latent representation of predicate $p$ in the RESCAL and Tree models, $\mathcal{G}_t$ represents the three-dimensional tensor latent representation of timestamp $t$ in the ConT model.}
  \label{fig:Tensor_models}
\end{figure*}

There are different ways for modeling the indicator function $f^{epi}(\cdot)$ or $f^{sem}(\cdot)$. In this paper, we will only investigate multilinear models derived from tensor decompositions and compositional operations. We now describe the models in detail. Graphical illustrations of the described models are shown in Figure~\ref{fig:Tensor_models}.

Table~\ref{tab:general_notation} and Table~\ref{tab:notation} summarize notations used throughout this paper for easy reference, while Table~\ref{tab:parameters} summarizes the number of parameters required for each model.\footnote{For DistMult, ComplEx, and HolE it is required that $\tilde{r} = \tilde{r}_t$. In our experiments (see Sections~\ref{sec:Episodic_exp} and~\ref{sec:Projection_exp}), in order to enable a fair comparison between the different models, we assume that the latent representations of entities, predicates, and time indices all have the same rank/dimensionality.}

\begin{table}[thp]
  \centering
  \caption{Summary of the general notations.}
  \label{tab:general_notation}
  \begin{tabular}{ll}
	\toprule
	\multicolumn{2}{c}{\textbf{General}} \\
	Symbol & Meaning\\
	\midrule
	$e_s$ & Entity for subject index $s$\\
	$e_o$ & Entity for object index $o$\\
	$e_p$ & Generalized entity for predicate index $p$\\
	$e_t$ & Generalized entity for time index $t$\\
	$\mathbf{a}_{e_i}$ & Latent representation of entity $e_i$\\
	$\mathbf{a}(e_{t_{start}})$ & Latent representation of starting timestamp\\
	$a_{e_i,r_i}$ & $r_i$-th element of $\mathbf{a}_{e_i}$\\
	$\tilde{r}$ & Rank/Dimensionality of $\mathbf{a}_{e_i}$ for $i \in \{s,p,o\}$\\
	$\tilde{r}_t$ & Rank/Dimensionality of $\mathbf{a}_{e_t}$\\
	$N_{e/p/t}$ & Number of entities / predicates / timestamps\\
	\bottomrule
  \end{tabular}
\end{table}

\begin{table}[thp]
  \centering
  \caption{Summary of the notations for semantic and episodic knowledge graphs.}
  \label{tab:notation}
  \begin{tabular}{llll}
	\toprule
	\multicolumn{2}{c}{\textbf{Semantic knowledge graphs}} &  \multicolumn{2}{c}{\textbf{Episodic knowledge graphs}} \\
	Symbol & Meaning & Symbol & Meaning\\
	\midrule
	$\chi$ & Sem. adjacency tensor & 	$\mathcal{E}$ & Epi. adjacency tensor\\
	$\Theta_\chi$ & Companion tensor of $\chi$ & $\Theta_\mathcal{E}$ & Companion tensor of $\mathcal{E}$\\
	$x_{s,p,o}$ & Value of ($e_s$, $e_p$, $e_o$) & $x_{t,s,p,o}$ & Value of ($e_t$, $e_s$, $e_p$, $e_o$)\\
	$\theta^{sem}_{s,p,o}$ & Logit of ($e_s$, $e_p$, $e_o$) & $\theta^{epi}_{t,s,p,o}$ & Logit of ($e_t$, $e_s$, $e_p$, $e_o$)\\
	$f^{sem}(\cdot)$ & Sem. indicator function & $f^{epi}(\cdot)$ & Epi. indicator function\\
	$\mathcal{G}^{sem}$ & Sem. core tensor & $\mathcal{G}^{epi}$ & Epi. core tensor\\
	$g^{sem}(\cdot)$ & Element of $\mathcal{G}^{sem}$ & $g^{epi}(\cdot)$ & Element of $\mathcal{G}^{epi}$\\
	\bottomrule
  \end{tabular}
\end{table}

\textbf{Tucker.} First, we consider the Tucker model for semantic tensor decomposition of the form $\theta^{sem}_{s, p, o} = \sum_{r_1, r_2, r_3=1}^{\tilde{r}} a_{e_s,r_1}a_{e_p,r_2}a_{e_o,r_3} g^{sem}(r_1,r_2,r_3)$.
Here, $g^{sem}(r_1, r_2, r_3)\in\mathbb{R}$ are elements of the core tensor $\mathcal{G}^{sem}\in\mathbb{R}^{\tilde{r}\times \tilde{r}\times\tilde{r}}$. Similarly, the indicator function of a four-way Tucker model for episodic tensor decomposition is of the form
\begin{align}
  &\theta^{epi}_{t, s, p, o} = \sum\limits_{r_1=1}^{\tilde{r}_t} \sum\limits_{r_2, r_3, r_4=1}^{\tilde{r}} \nonumber\\
  &a_{e_t,r_1} a_{e_s,r_2} a_{e_p,r_3} a_{e_o,r_4} g^{epi}(r_1,r_2,r_3, r_4),
\end{align}
with a four dimensional core tensor $\mathcal{G}^{epi}\in\mathbb{R}^{\tilde{r}_t\times\tilde{r}\times\tilde{r}\times\tilde{r}}$.
Note that this is a constraint Tucker model, since, as in RESCAL, entities have unique representations, independent of the roles as subject or object.

\textbf{RESCAL.} Another model closely related to the semantic Tucker tensor decomposition is the RESCAL model, which has shown excellent performance in modelling KGs \cite{nickel2011three}. In RESCAL, subjects and objects have vector latent representations, while predicates have matrix latent representations. The indicator function of RESCAL for modeling semantic KGs takes the form $\theta^{sem}_{s,p,o} = \sum_{r_1, r_2=1}^{\tilde{r}} a_{e_s,r_1}g_p(r_1,r_2)a_{e_o,r_2}$, where $g_p(r_1, r_2)$ represents the matrix latent representation for the predicate $e_p$. Then next two models,  Tree and ConT,  are novel  generalizations of RESCAL to episodic tensors.

\textbf{Tree.} From a practical perspective, training an episodic Tucker tensor model is very expensive since the computational complexity is approximately $\tilde{r}^4$. Tensor networks provide a general and flexible framework to design nonstandard tensor decompositions \cite{cichocki2014era,cichocki2014tensor}. One of the simplest tensor networks is a tree tensor decomposition ($\mathcal{T}$) of the episodic indicator function, which is illustrated in  compositional operations. We now describe the models in detail. Graphical illustrations of the described models are shown in Figure~\ref{fig:Tensor_models}(e). Therefore, we propose a tree tensor decomposition ($\mathcal{T}$) of the episodic indicator function. The tree $\mathcal{T}$ is partitioned into two subtrees $\mathcal{T}_1$ and $\mathcal{T}_2$, wherein subject $e_s$ and time $e_t$ reside in $\mathcal{T}_1$, while object $e_o$ and an auxiliary time $e_t$ reside in $\mathcal{T}_2$. $\mathcal{T}_1$ and $\mathcal{T}_2$ are connected with $e_p$ through two core tensors $\mathcal{G}_1$ and $\mathcal{G}_2$. Thus, the indicator function can be written as
\begin{align}
  &\theta^{epi}_{t, s, p, o} =\nonumber \sum\limits_{r_1, r_6 =1}^{\tilde{r}_t} \sum\limits_{r_2, r_3, r_4, r_5=1}^{\tilde{r}}\\
  &a_{e_t, r_1}a_{e_s, r_2}g_1(r_1, r_2, r_3)g_p(r_3, r_4)g_2(r_4, r_5, r_6) a_{e_o, r_5} a_{e_t, r_6}.
\end{align}
Within $\mathcal{T}$, we reduce the four-way core tensor in Tucker into two three-dimensional tensors $\mathcal{G}_1$ and $\mathcal{G}_2$, so that the computational complexity of $\mathcal{T}$ is approximately $\tilde{r}^3$.

\textbf{ConT.} ConT is another generalization of the RESCAL model to episodic tensors with reduced computational complexity of approximately $\tilde{r}^3$. The idea is that another way of reducing the complexity is by contracting indices of the core tensor. Therefore, we contract the $\mathcal{G}$ from Tucker with the time index giving a three-way core tensor $\mathcal{G}_t$ for each time instance. The indicator function takes the form
\begin{equation}
  \theta^{epi}_{t, s, p, o} = \sum\limits_{r_1, r_2, r_3 =1}^{\tilde{r}} a_{e_s, r_1} a_{e_p, r_2} a_{e_o, r_3} g_t(r_1, r_2, r_3).
\end{equation}
In this model, the tensor $\mathcal{G}_t$ resembles the relation-specific matrix $\mathcal{G}_p$ from RESCAL. Later, we will see that ConT is a superior model for modeling episodic knowledge graphs due to the representational flexibility of its high-dimensional tensor $\mathcal{G}_t$ for the time index.

Even though the complexity of Tree and ConT is reduced as compared to episodic Tucker, the three-dimensional core tensor might cause rapid overfitting during training. Therefore, we next propose episodic generalization of compositional models, such as DistMult~\cite{yang2014embedding}, HolE~\cite{nickel2016holographic} and ComplEx~\cite{trouillon2016complex}. For those models, the number of parameters only increases linearly with the rank.

\textbf{DistMult.} DistMult~\cite{yang2014embedding} is a simple generalization of the CP model, by enforcing the constraint that entities should have unique representations. Episodic DistMult   takes the form $\theta^{epi}_{t, s, p, o} = \sum_{i =1}^{\tilde{r}}\lambda_i a_{e_t, i} a_{e_s, i} a_{e_p, i} a_{e_o, i}$. Here, we require that vector latent representations of entities, predicates, and timestamps have the same rank. DistMult is a special case of Tucker having a core tensor with only diagonal elements $\lambda_i$.

\textbf{HolE.} Holographic embedding (HolE)~\cite{nickel2016holographic} is a state-of-art link prediction and knowledge graph completion method, which is inspired by holographic models of associative memory.

HolE uses circular correlation to generate a compositional representation from inputs $e_s$ and $e_o$. The indicator of HolE reads $ \theta^{sem}_{s, p, o} = \textbf{a}_{e_p} \cdot (\textbf{a}_{e_s} \star \textbf{a}_{e_o})$, where $\star: \mathbb{R}^d \times \mathbb{R}^d \rightarrow \mathbb{R}^d$ denotes the circular correlation $[\textbf{a} \star \textbf{b}]_k = \sum_{i=0}^{d-1}a_ib_{(k+i) \operatorname{mod} d}$.
We define the episodic extension of HolE as
\begin{equation}
   \theta^{epi}_{t, s, p, o} = \textbf{a}_{e_t} \cdot \left(\textbf{a}_{e_p}\star\left(\textbf{a}_{e_s} \star \textbf{a}_{e_o} \right)\right).
   \label{epiHolE}
\end{equation}

As argued by \cite{nickel2016holographic}, HolE employs a holographic reduced representation \cite{plate1995holographic} to store and retrieve the predicates from $e_s$ and $e_o$. Analogously, episodic HolE should be able to retrieve the stored timestamps from $e_p$, $e_s$ and $e_o$. In the semantic case, $e_p$ can be retrieved if existing triple relations are stored via circular convolution $\ast$, and superposition in the representation $\textbf{a}_{e_o} =\sum_{(s,p) \in \mathcal{S}_{o}} \textbf{a}_{e_p} \ast \textbf{a}_{e_s}$, where $\mathcal{S}_o$ is the set of all true triples given $e_o$. This is based on the fact that $\textbf{a}\star\textbf{a}\approx\delta$ \cite{nickel2016holographic}. Analogously, the stored timestamp $e_t$ for an event can be retrieved if all existing episodic events are stored via $\ast$, and superposition in the representation of $e_o$,  $\textbf{a}_{e_o}=\sum_{(t,s,p)\in\mathcal{S}_o}\textbf{a}_{e_t} \ast (\textbf{a}_{e_p} \ast \textbf{a}_{e_s})$, where $\mathcal{S}_o$ is the set of all true quadruples $(t,s,p,o)$ given $e_o$. However, high order circular correlation/convolution will increase the inaccuracy of retrieval. Another motivation for our episodic extension~\eqref{epiHolE} is that a compositional operator of the form $\textbf{a}_{e_t} \cdot \tilde{f}$ allows a projection from episodic memory to semantic memory, to be detailed later.

\textbf{ComplEx.} Complex embedding (ComplEx)~\cite{trouillon2016complex} is another state-of-art method closely related to HolE. It  can accurately describe both symmetric and antisymmetric relations.  HolE is a special case of ComplEx with imposed conjugate symmetry on embeddings~\cite{DBLP:journals/corr/HayashiS17}. Thus, ComplEx has more degrees of freedom, if compared to   HolE.
For the semantic complex embedding, the indicator function is $\theta^{sem}_{s,p,o}=\operatorname{Re}\left(\sum_i^{\tilde{r}} a_{e_s, i} a_{e_p, i}, \bar{a}_{e_o, i}\right)$ with complex valued $\textbf{a}$ and where the bar indicates the complex conjugate. To be consistent with the episodic HolE, the episodic complex embedding is defined as\footnote{One can show that Eq.~\eqref{epiHolE} is equivalent to Eq.~\eqref{epiComplEx} by converting it to the frequency domain~\cite{DBLP:journals/corr/HayashiS17}. Then, $\theta^{epi}_{t,s,p,o} \propto \boldsymbol{\omega}_{e_t}^T (\bar{\boldsymbol{\omega}}_{e_p} \odot \bar{\boldsymbol{\omega}}_{e_s} \odot \boldsymbol{\omega}_{e_o})$, where $\boldsymbol{\omega}=\mathscr{F}(\textbf{a})\in\mathbb{C}^{\tilde{r}}$ are the discrete Fourier transforms of embeddings $\textbf{a}$, and using the fact that $\boldsymbol{\omega}$ is conjugate symmetric for real vector $\textbf{a}$.}
\begin{equation}
  \theta^{epi}_{t, s,p,o}=\operatorname{Re}\left(\sum_i^{\tilde{r}} a_{e_t, i}  a_{e_s, i} a_{e_p, i}, \bar{a}_{e_o, i}\right).
  \label{epiComplEx}
\end{equation}

\section{Experiments on Episodic Models}
\label{sec:Episodic_exp}

We investigate the proposed tensor and compositional models with experiments which are evaluated on two datasets:

\textbf{ICEWS.} The Integrated Conflict Early Warning System (ICEWS) dataset \cite{ward2013comparing} is a natural episodic dataset recording dyadic events between different countries. An example entry could be (\emph{Turkey}, \emph{Syria}, \emph{Fight}, \emph{12/25/2014}). These dyadic events are aggregated into a four-way tensor $\mathcal{E}$ with $258$ entities, $20$ relation types, and $72$ timestamps, which has in total $320,118$ positive $(e_t, e_s, e_p, e_o)$ quadruples~\footnote{Note that for an episodic event the dataset contains all the quadruples $(e_{t_i}, e_s, e_p, e_0)$ for $t_i \in \{ t_{start}, t_{start} +1, \cdots, t_{end} -1, t_{end}\}$. }. This dataset was first created and used in \cite{schein2015bayesian}.
From this ICEWS dataset, a semantic tensor is generated by extracting consecutive events that last until the last timestamp, constituting the \emph{current}~\footnote{\emph{Current} always indicates the last timestamp/timestamps of the applied episodic KGs.} semantic facts of the world.

\textbf{GDELT.} The Global Database of Events, Language and Tone (GDELT) \cite{ward2013comparing} monitors the world's news media in broadcast, print and web formats from all over the world, daily since January 1, 1979 \footnote{\url{https://www.gdeltproject.org/about.html}}. We use GDELT as a large episodic dataset. For our experiments, GDELT data is collected from January 1, 2012 to December 31, 2012 (with a temporal granularity of 24 hrs). These events are aggregated into an episodic tensor $\mathcal{E}$ with $1100$ entities, $180$ relation types, and $366$ timestamps, which has in total $2,563,561$ positive $(e_t, e_s, e_p, e_o)$ quadruples.

We assess the quality of episodic information retrieval on both datasets for the proposed tensor and compositional models. Since both episodic datasets only consist of positive quadruples, we generated negative episodic instances following the protocol of corrupting semantic triples given by Bordes \cite{bordes2013translating}: negative instances of an episodic quadruple $(e_s, e_p, e_o, e_t)$ are drawn by corrupting the object $e_o$ to $e_{o'}$ or the timestamp $e_t$ to $e_{t'}$ , meaning that $(e_s,e_p, e_{o'}, e_t)$ serves as a negative evidence of the episodic event at time instance $e_t$, and $(e_s, e_p, e_o, e_{t'})$ is a true fact which cannot be correctly recalled at time instance $e_{t'}$. During training, for each positive sample in a batch we assigned two negative samples with corrupted object or corrupted subject.

\begin{table*}[t]

  \setlength{\tabcolsep}{0.15pc}
  \footnotesize

  \centering
  \caption{Number of parameters for different models and the runtime of one training epoch on the GDELT dataset.}
  \label{tab:parameters}
  \begin{tabular}{l l l c r r r}
	\toprule
	& & & & \multicolumn{3}{c}{Runtime} \\
	\cmidrule{5-7}
	Model & Semantic  & Episodic & Complexity & rank $40$ & rank $60$ & rank $150$ \\
    \midrule
    DistMult     & $(N_e+N_p+1)\tilde{r}$  & $(N_e+N_p+N_t+1)\tilde{r}$  & $\mathcal{O}(\tilde{r})$ & $35.2s$ & $36.4s$  & $53.7s$ \\
	HolE          & $(N_e+N_p)\tilde{r}$   &  $(N_e+N_p)\tilde{r}$  & $\mathcal{O}(\tilde{r}\log\tilde{r})$ & $42.8s$  & $43.2s$  & $59.0s$  \\
	ComplEx      & $2(N_e+N_p)\tilde{r}$  & $2(N_e+N_p+N_t)\tilde{r}$  & $\mathcal{O}(\tilde{r})$ & $40.1s$ & $42.4s$  & $57.5s$ \\
	\midrule
	Tree          & $-$                         & $N_e\tilde{r}+N_p\tilde{r}^2+(N_t+2\tilde{r}^2)\tilde{r}_t$
	              & $\mathcal{O}(\tilde{r}^3)$  & $133.6s$  & $160.2s$   & $-$  \\
	ConT          & $-$                         & $(N_e+N_p)\tilde{r}+N_t\tilde{r}^3$
	              & $\mathcal{O}(\tilde{r}^3)$  & $95.4s$   & $226.1s$   & $-$  \\
	Tucker        & $(N_e+N_p)\tilde{r}+\tilde{r}^3$  & $(N_e+N_p) \tilde{r}+(N_t+\tilde{r}^3)\tilde{r}_t$
	              & $\mathcal{O}(\tilde{r}^4)$  & $144.2s$  & $387.9s$   & $-$  \\
    \bottomrule
  \end{tabular}
\end{table*}

The model performance is evaluated using the following scores. To retrieve the occurrence time, for each true quadruple, we replace the time index $e_t$ with every other possible time index $e_{t'}$, compute the value of the indicator function $\theta^{epi}_{t', s, p, o}$, and rank them in a decreasing order. We filter the ranking as in \cite{bordes2013translating} by removing all quadruples where $x_{t',s,p,o}=1$ and $t \neq t'$, in order to eliminate ambiguity during episodic information retrieval. Similarly, we evaluated the retrieval of the predicate between a given subject and object at a certain time instance by computing and ranking the indicator $\theta^{epi}_{t, s,p',o}$. We also evaluated the retrieval of entities by ranking and averaging the filtered indicators $\theta_{t,s',p,o}$ and $\theta_{t,s,p,o'}$. To measure the generalization ability of the models, we report different measures of the ranking: mean reciprocal rank (MRR), and Hits@n on the test dataset.

The datasets were split into train, validation, and test sets that contain the most frequently appearing entities in the episodic knowledge graphs. Training was performed by minimizing the logistic loss~\eqref{logisticloss}, and was terminated using early stopping on the validation dataset by monitoring the filtered MRR recall scores every $\{50, 100\}$ epochs depending on the models, where the maximum training duration was $500$ epochs. This ensures that the generalization ability of unique latent representations of entities doesn't suffer from overfitting. Before training, all model parameters are initialized using Xavier initialization \cite{glorot2010understanding}. We also apply an $l2$ norm penalty on all parameters for regularization purposes (see Eq.~\eqref{logisticloss}).

In Table~\ref{tab:parameters} we summarize the runtime for one training epoch on the GDELT dataset for different models at ranks $\tilde{r}=\tilde{r}_t \in\{40, 60, 150\}$. All experiments were performed on a single Tesla K$80$ GPU. In the following experiments, for compositional models we search rank in $\{100, 150\}$, while for tensor models we search optimal rank in $\{40, 50, 60\}$ since larger ranks could lead to overfitting rapidly. Loss function is minimized with Adam method \cite{kingma2014adam} with the learning rate selected from $\{0.001, 1e-4, 5e-5\}$.

\begin{table*}[t]
  \setlength{\tabcolsep}{0.2pc}
  \small
  \caption{Filtered results of inferring missing entities and predicates of episodic quadruples evaluated on the GDELT dataset.}
  \label{table:Gdelt_results}
  \centering
  \begin{tabular}{lccccccccc}
    \toprule
    & \multicolumn{4}{c}{Entity} & & \multicolumn{4}{c}{Predicate} \\
    \cmidrule{2-5} \cmidrule{7-10}
    Method & MRR & @1 & @3 & @10   &&   MRR & @1 & @3 & @10 \\
    \midrule
    DistMult  & 0.182 & 6.55   & 19.77  & 43.70  &&  0.269  &  12.65  &  30.29  & 59.40 \\
    HolE      & 0.177 & 6.67   & 18.95  & 41.84  &&  0.256  &  11.81  &  28.35  & 57.73 \\
    ComplEx   & 0.172 & 6.54   & 17.52  & 41.56  &&  0.255  &  12.05  &  27.75  & 56.60 \\
    Tree      & 0.196 & 8.17   & 21.00  & 44.65  &&  0.274  &  \textbf{13.30}  &  30.66  & 60.05 \\
    Tucker    & 0.204 & 8.93   & 21.85  & \textbf{46.35}  &&  \textbf{0.275}  &  12.69  &  \textbf{31.35}  & \textbf{60.70} \\
    ConT      & \textbf{0.233} & \textbf{13.85}  & \textbf{24.65}  & 42.96  &&  0.263  &  12.83  &  29.27  & 57.30 \\   %
    \bottomrule
  \end{tabular}
\end{table*}
\normalsize

We first assess the filtered MRR, Hits@1, Hits@3, and Hits@10 scores of inferring missing entities and predicates on the GDELT test dataset. Table~\ref{table:Gdelt_results} summarizes the results. Generalizations on the test dataset indicate the inductive reasoning capability of the proposed models. This generalization can be useful for the completion of evolving KGs with missing records, such as clinical datasets. It can be seen that tensor models are able to outperform compositional models consistently on both entity and predicate prediction tasks. ConT has the best inference results on the entity-related tasks, while Tucker performs better on the predicate-related tasks. The superior Hits@1 result of ConT on the entity prediction indicates that there are easily to be fitted entities in the GDELT dataset along the timestamps. In fact, the GDELT dataset is unbalanced, and episodic quadruples related to certain entities dominate in the episodic Knowledge graph, such as quadruples containing the entities \emph{USA}, or \emph{UN}. Experiment results on balanced and extremely sparse episodic dataset will be reported in the following.

\begin{table*}[t]
  \setlength{\tabcolsep}{0.2pc}
  \small
  \caption{Filtered results for entities and predicates recollection/prediction evaluated on the ICEWS dataset.}
  \label{table:Icews_results}
  \centering
  \begin{tabular}{lccccccccc}
    \toprule
    & \multicolumn{4}{c}{Entity} & & \multicolumn{4}{c}{Predicate} \\
    \cmidrule{2-5} \cmidrule{7-10}
    Method & MRR & @1 & @3 & @10   &&   MRR & @1 & @3 & @10 \\
    \midrule
    DistMult  & 0.222 & 9.72   & 22.48  & 52.32  &&  0.520  &  33.73  &  62.25  & 91.13 \\
    HolE      & 0.229 & 9.85   & 23.49  & 54.21  &&  0.517  &  31.55  &  65.47  & 93.59 \\
    ComplEx   & 0.229 & 8.94   & 23.53  & \textbf{57.72}  &&  0.506  &  30.99  &  61.46  & 93.44 \\
    Tree      & 0.205 & 10.48  & 19.84  & 42.81  &&  0.554  &  36.62  &  67.25  & 94.70 \\
    Tucker    & 0.257 & 12.88  & 27.10  & 54.43  &&  \textbf{0.563}  &  36.96  &  \textbf{69.55}  & \textbf{95.43} \\
    ConT      & \textbf{0.264} & \textbf{15.71}  & \textbf{29.60}  & 46.67  &&  0.557  &  \textbf{38.12}  &  67.76  & 87.71 \\
    \bottomrule
  \end{tabular}
\end{table*}
\normalsize

Next, Table~\ref{table:Icews_results} shows the MRR, Hits@1, Hits@3, and Hits@10 scores of inferring missing entities and predicates on the ICEWS test dataset. Similarly, we can read that tensor models outperform compositional models on both missing entity and predicate inference tasks. The superior Hits@1 result of ConT for the missing entity prediction indicates again that the ICEWS dataset is unbalanced, and episodic quadruples related to certain entities dominate.

\begin{table}[htp!]

  \setlength{\tabcolsep}{0.2pc}
  \small

  \caption{Filtered recall scores for entities and timestamps recollection on the ICEWS (rare) training dataset.}
  \label{table:Icews_rare_results}
  \centering
  \begin{tabular}{ll*{5}{c}}
    \toprule
    & & \multicolumn{2}{c}{Timestamp} &  & \multicolumn{2}{c}{Entity} \\
    \cmidrule{3-4} \cmidrule{6-7}
    Method & Rank & MRR  & @3 &  & MRR & @3 \\
    \midrule
	DistMult   &  200  &  0.257  &  27.0   & &     0.211   &  21.9  \\
	HolE       &  200  &  0.216  &  20.8   & &     0.179   &  16.3  \\
	ComplEx    &  200  &  0.354  &  40.3   & &     0.301   &  33.2  \\
	Tree       &  40   &  0.421  &  55.3   & &     0.314   &  35.7  \\
	Tucker     &  40   &  0.923  &  98.9   & &     0.893   &  97.1  \\
	\midrule
	ConT       &  40   & \textbf{0.982} & \textbf{99.7}   & &    \textbf{0.950} &  \textbf{97.9} \\
    \bottomrule
  \end{tabular}
\end{table}
\normalsize

\begin{figure}[htp]
\begin{center}
  \includegraphics[width=0.48\linewidth]{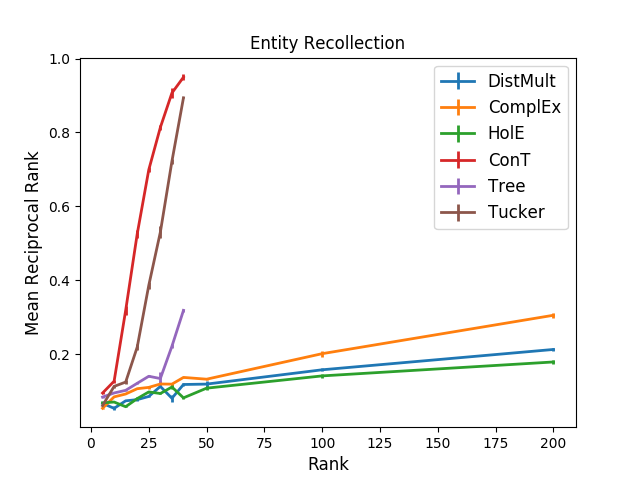}
  \includegraphics[width=0.48\linewidth]{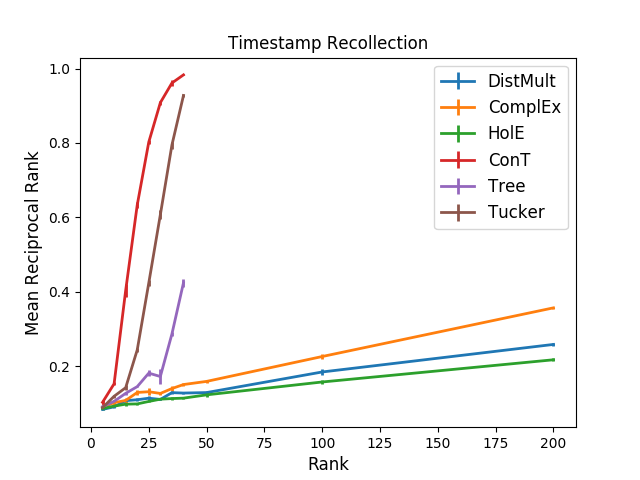}
  \caption{Filtered MRR scores vs. rank for the entities (left) and timestamps (right) recollection on the ICEWS (rare) training dataset.}
  \label{fig:icews rare}
\end{center}
\end{figure}

The recollection of the exact occurrence time of a significant past event (e.g. unusual, novel, attached with emotion) is also an important capability of episodic cognitive memory function. In order to manifest this perspective of proposed models, Table~\ref{table:Icews_rare_results} shows the filtered MRR, and Hits@3 scores for the timestamps and entities recollection on the episodic ICEWS (rare) training dataset, where rank column registers the optimal and minimum rank $\tilde{r}=\tilde{r}_t$ having the outstanding recall scores. Figure~\ref{fig:icews rare} further displays the filtered MRR score as a function of rank.
Unlike the original ICEWS, which contains many consecutive events that last from the first to the last timestamp leading to unreasonably high filtered timestamp recall scores, this ICEWS (rare) dataset consists of rare temporal events that happen less than three times throughout the whole time and starting points of events.

The outstanding performance of ConT compared with other compositional models indicates the importance of large dimensionality of time latent representation for the episodic tensor reconstruction / episodic memory recollection. Recall that for ConT the real \textit{dimension} of the latent representation of time is actually $\tilde{r}^3$ after flattening $\mathcal{G}_t$. This flexible latent representation for time could compress almost all the semantic triples that occur at a certain instance \footnote{This observation has its biological counterpart. In fact, the entorhinal cortex, which plays an important role in the formation of episodic memory, is the main part of the adult hippocampus that shows neurogenesis \cite{deng2010new}. In an adult human, approximately 700 new neurons are added per day through hippocampal neurogenesis, which are believed to perform sensory and spatial information encoding, as well as temporal separation of events \cite{lazarov2016hippocampal,spalding2013dynamics}.}.

\section{Semantic Memory from Episodic Memory with Marginalization}
\label{sec:Projection_exp}

We already discussed that a semantic KG might be related to a human semantic memory and that an episodic KG might be related to a human episodic memory. It has been speculated that episodic and semantic memory must be closely related, and that semantic memory is generated from episodic memory by some training process \cite{mcclelland1995there,nadel2000multiple}. As a very simple implementation of that idea, we propose that a semantic memory could be generated from episodic memory by marginalizing time. Thus, both types of memories would rely on identical representations and the marginalization step can be easily performed: Since probabilistic tensor models belong to the classes of sum-product nets, a marginalization simply means an integration over all time representations.

Thus, in the  second set of experiments, we test the  hypothesis that semantic memory can be derived from episodic memory by projection. In other words, a semantic knowledge graph containing \emph{current} semantic facts can be approximately constructed after modeling a corresponding episodic knowledge graph via marginalization. A  marginalization  can be  performed by activating all time index neurons, i.e., summing over all $\textbf{a}_{e_t}$, since, e.g., Tucker decompositions are an instance of a so-called sum-product network \cite{poon2011sum}. However, events having start as well as end timestamps cannot simply be integrated into our \emph{current} semantic knowledge describing what we \emph{know} now. For example, (Ban Ki-moon, SecretaryOf, UN) is not consistent with what we \emph{know} currently. To resolve this problem, we introduce two types of time indices, $e_{t_{start}}$ and $e_{t_{end}}$, having the latent representations $\textbf{a}(e_{t_{start}})$ and $\textbf{a}(e_{t_{end}})$, respectively. Those time indices can be used to construct the episodic tensor $\mathcal{E}_{start}$ aggregating the start timestamps of consecutive events, as well as the episodic tensor $\mathcal{E}_{end}$ aggregating the end timestamps\footnote{E.g., if the duration of a triple event $(e_s,e_p,e_o)$ lasts from $t_{start}$ to $t_{end}$, the quadruple $(e_s,e_p,e_o, e_{ t_{start}})$ is stored in $\mathcal{E}_{start}$, while $(e_s, e_p, e_o, e_{t_{end}})$ is stored $\mathcal{E}_{end}$ only if $t_{end}< T$ (where $T$ is the last timestamp). In other words, events that last until the last timestamp do not possess $e_{end}$.}.

For the projection, instead of only summing over $\textbf{a}(e_{t_{start}})$, we also subtract the sum over $\textbf{a}(e_{t_{end}})$. In this way, we can achieve the effect that events that have terminated already (i.e., have an end time index smaller than the current time index) are not integrated into the current semantic facts. Now, to test our hypothesis that this extended projection allows us to derive semantic memory from episodic memory, we trained HolE, DistMult, ComplEx, ConT, and Tucker on the episodic tensors $\mathcal{E}_{start}$ and $\mathcal{E}_{end}$ as well as on the semantic tensor $\chi$ derived from ICEWS. Note that only these models allow projection, since their indicator functions can be written in the form $\theta^{epi}_{t,s,p,o}=\textbf{a}_{e_t}\cdot\tilde{f}$, where $\tilde{f}$ can be arbitrary function of $\textbf{a}_{e_s}$, $\textbf{a}_{e_p}$, and $\textbf{a}_{e_o}$ depending on the model choice\footnote{For ConT, $\theta^{epi}_{t,s,p,o} = \emph{flatten}(g_t) \cdot (\textbf{a}_{e_s}\otimes \textbf{a}_{e_p} \otimes \textbf{a}_{e_o})$, where $\otimes$ denotes the outer product. For ComplEx, $\theta^{epi}_{t,s,p,o}=\operatorname{Re}(\textbf{a}_{e_t}) \cdot \operatorname{Re} (\textbf{a}_{e_s} \odot \textbf{a}_{e_p} \odot \bar{\textbf{a}}_{e_o}) - \operatorname{Im}(\textbf{a}_{e_t}) \cdot \operatorname{Im} (\textbf{a}_{e_s} \odot \textbf{a}_{e_p} \odot \bar{\textbf{a}}_{e_o})$, where $\odot$ denotes the Hadamard product. The Tree model cannot be written in this form since $e_t$ resides in both subtrees $\mathcal{T}_1$ and $\mathcal{T}_2$.}. The model parameters are optimized using the margin-based ranking loss~\eqref{rankingloss}\footnote{For the projection experiment, we omit the sigmoid function in Eq.~\eqref{rankingloss}, train and interpret the multilinear indicator $\theta^{epi}_{t,s,p,o}= \textbf{a}_{e_t}\cdot \tilde{f}(\textbf{a}_{e_s}, \textbf{a}_{e_p}, \textbf{a}_{e_o})$ directly as the probability of episodic quadruple. Only in this way of training, a projection is mathematically legitimate.}.

Training was first performed on the episodic tensor $\mathcal{E}_{start}$, and then on $\mathcal{E}_{end}$ with \emph{fixed} $\textbf{a}_{e_s}$, $\textbf{a}_{e_p}$, and $\textbf{a}_{e_o}$ obtained from the training on $\mathcal{E}_{start}$, since we assume that latent representations for subject, object, and predicate of a consecutive event do not change during the event. Note that after training in this way, we could recall the starting and terminal point of a consecutive event (see the episodic tensor reconstruction experiments in Section~\ref{sec:Episodic_exp}), or infer a \emph{current} semantic fact solely from the latent representations instead of rule-based reasoning.

\begin{figure*}[htp]
  \centering
  \includegraphics[width=0.4\textwidth]{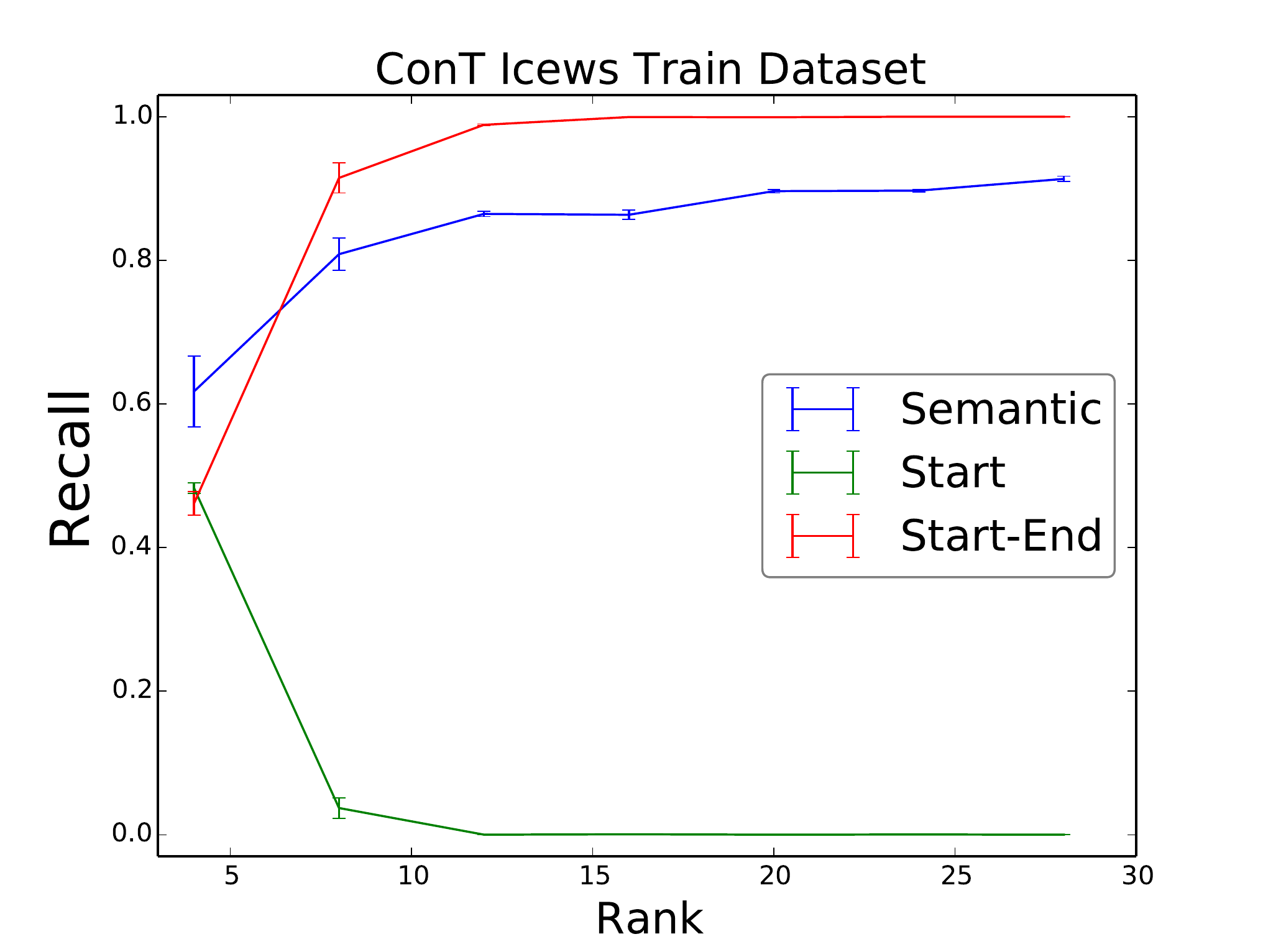}
  \includegraphics[width=0.4\textwidth]{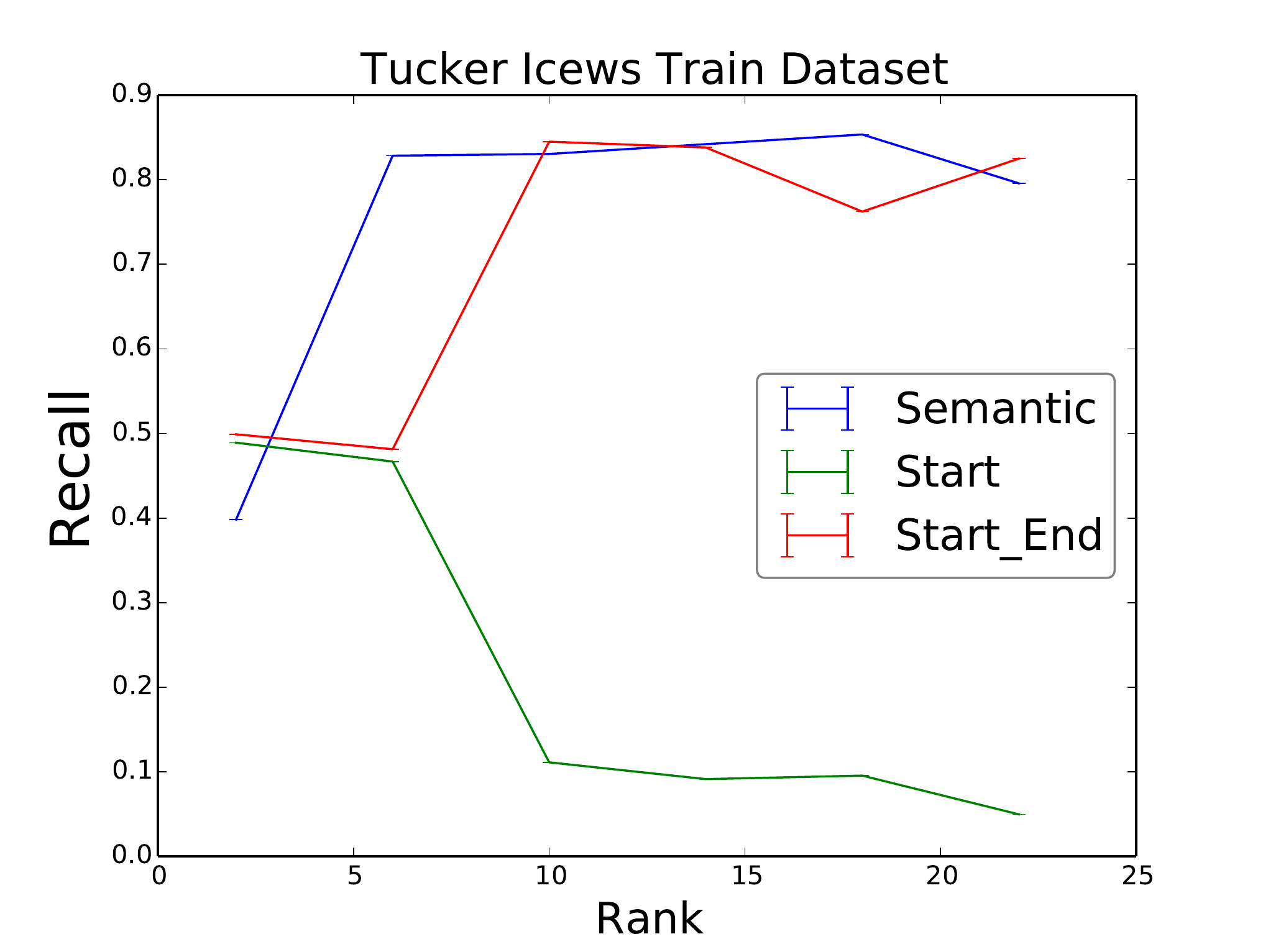}
  \includegraphics[width=0.4\textwidth]{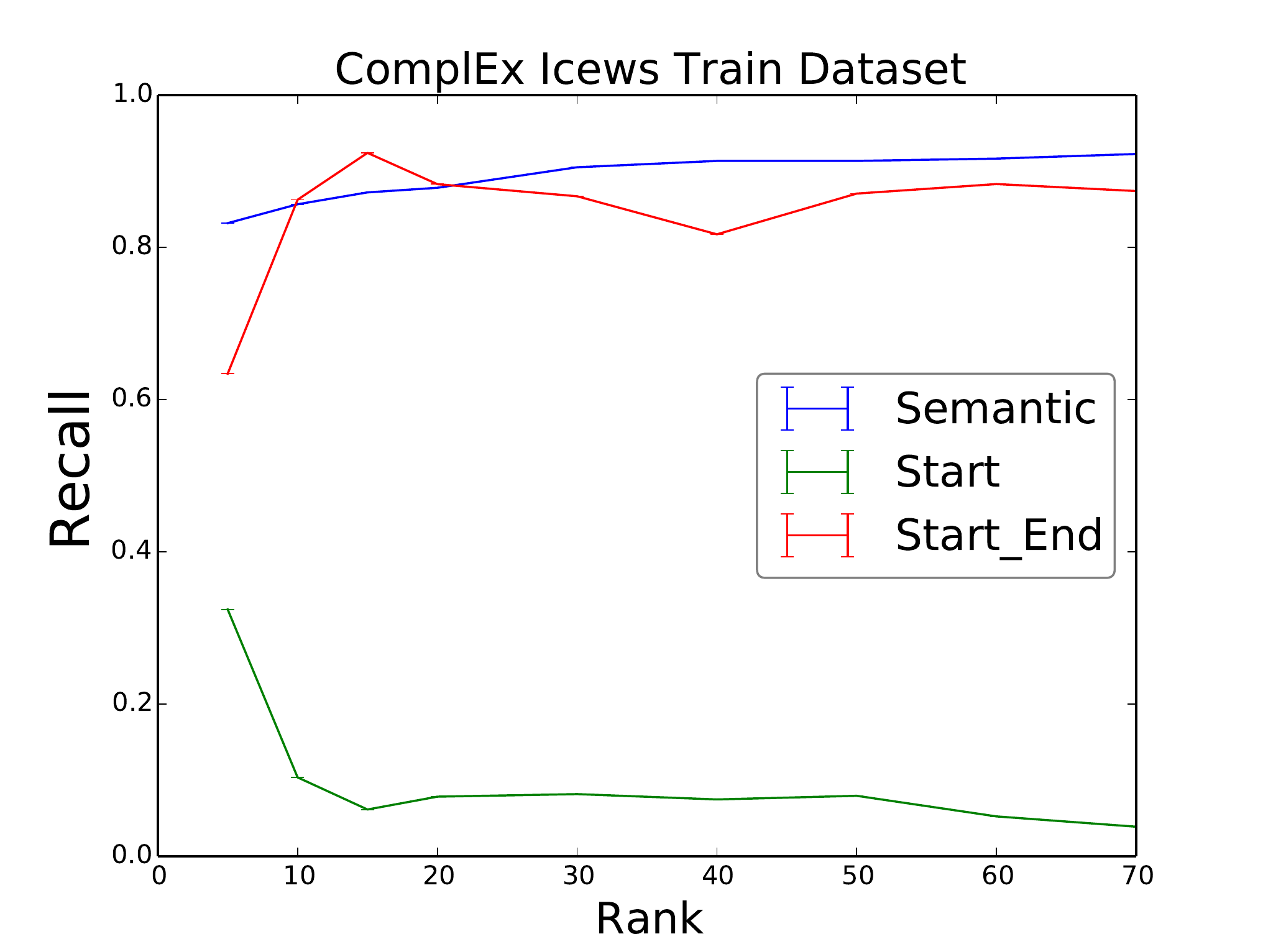}
  \includegraphics[width=0.4\textwidth]{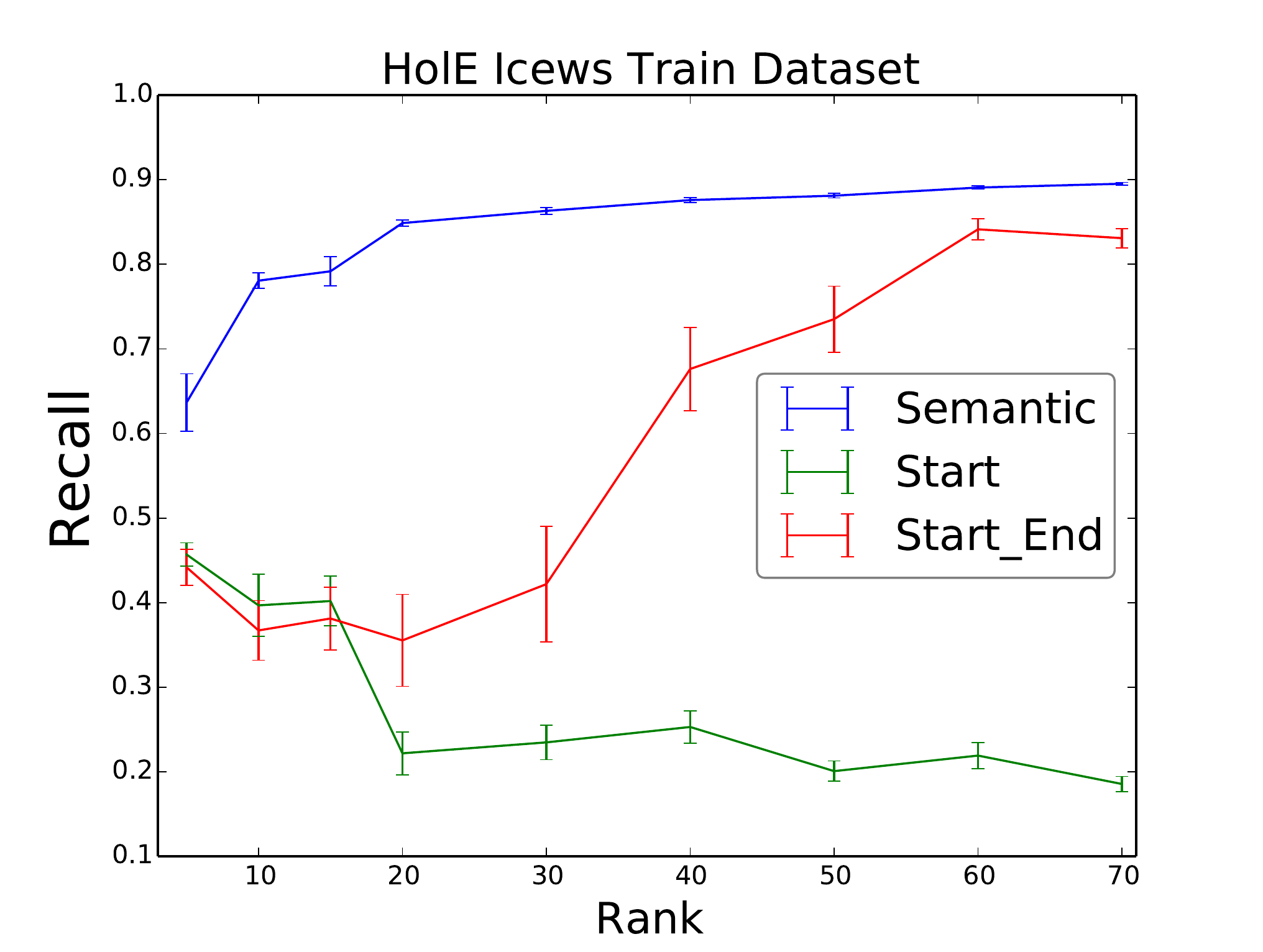}
  \caption{\small Recall scores vs. rank for the episodic-to-semantic projection on the ICEWS dataset with two different projection methods.}
  \label{fig:Projection_recall}
\end{figure*}

\begin{figure*}[htp]
  \centering
  \includegraphics[width=0.4\textwidth]{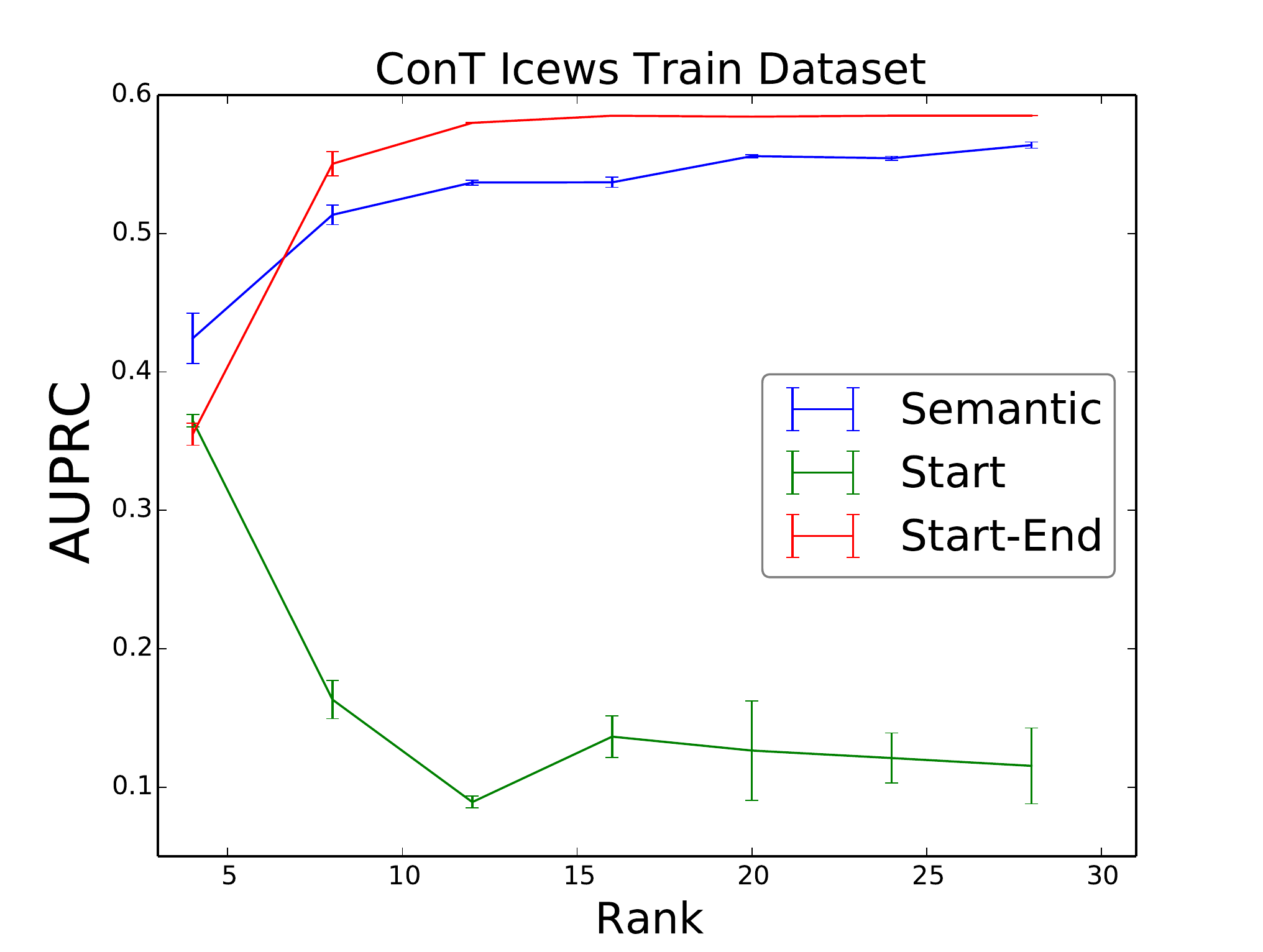}
  \includegraphics[width=0.4\textwidth]{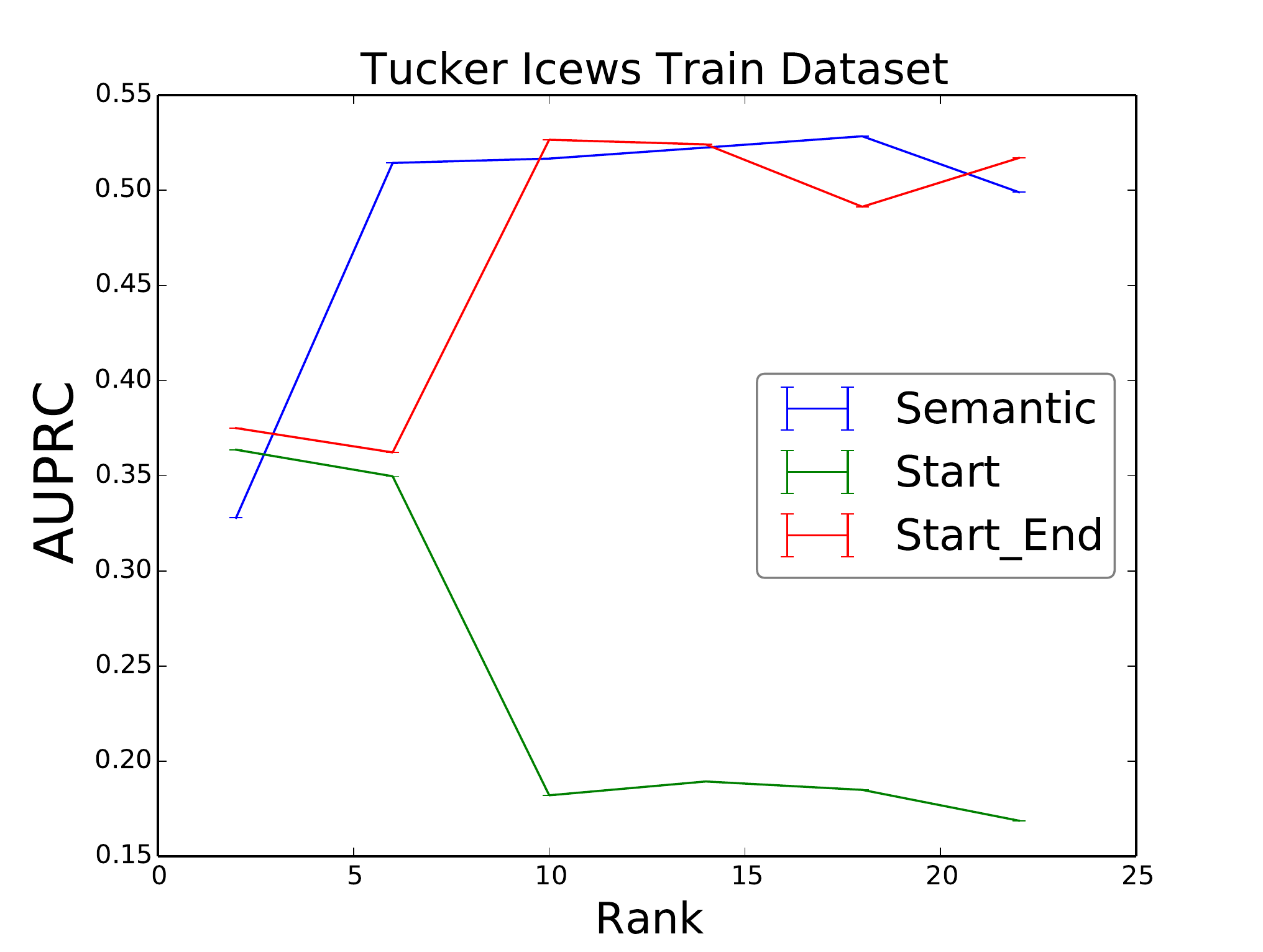}
  \includegraphics[width=0.4\textwidth]{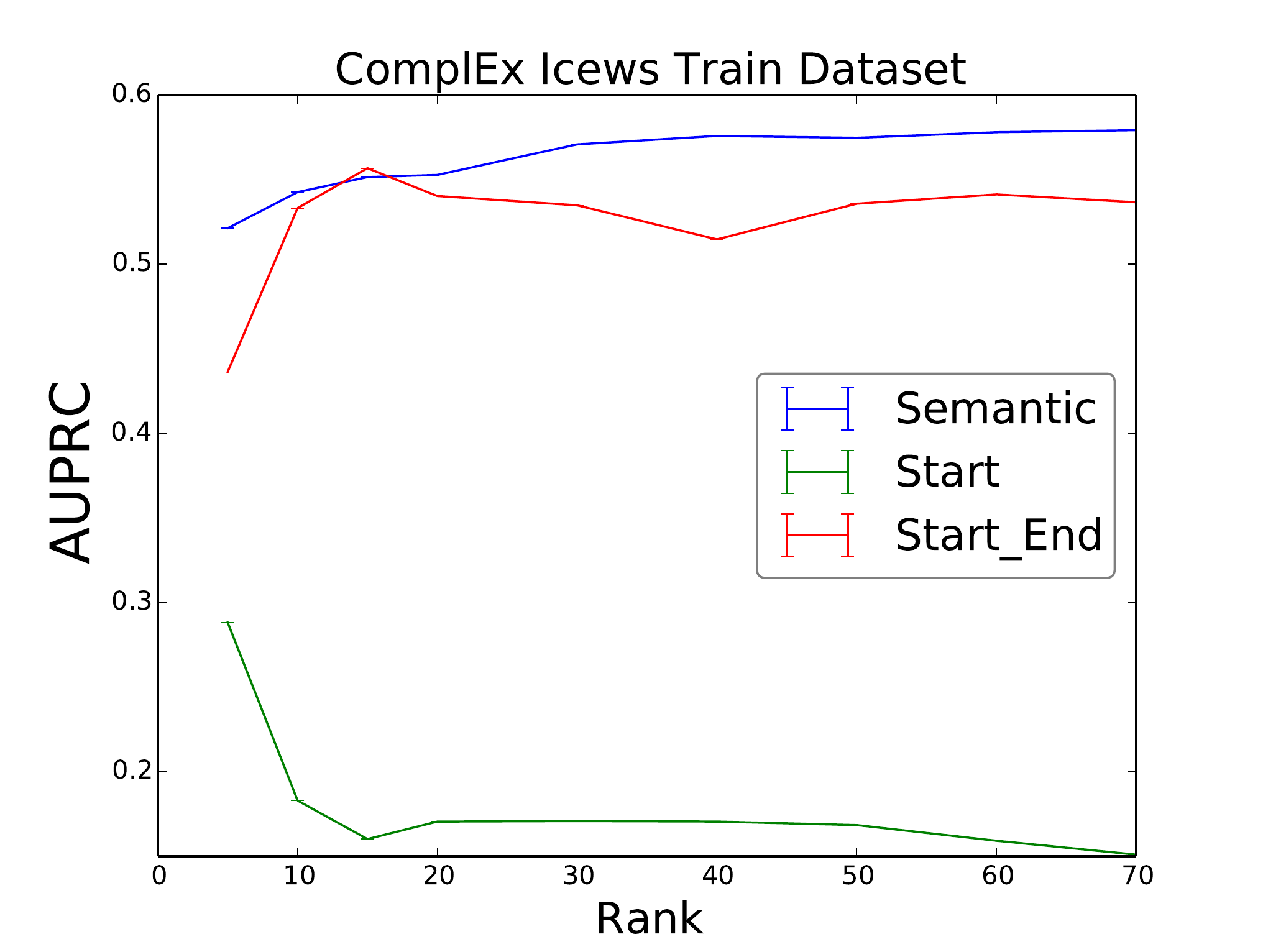}
  \includegraphics[width=0.4\textwidth]{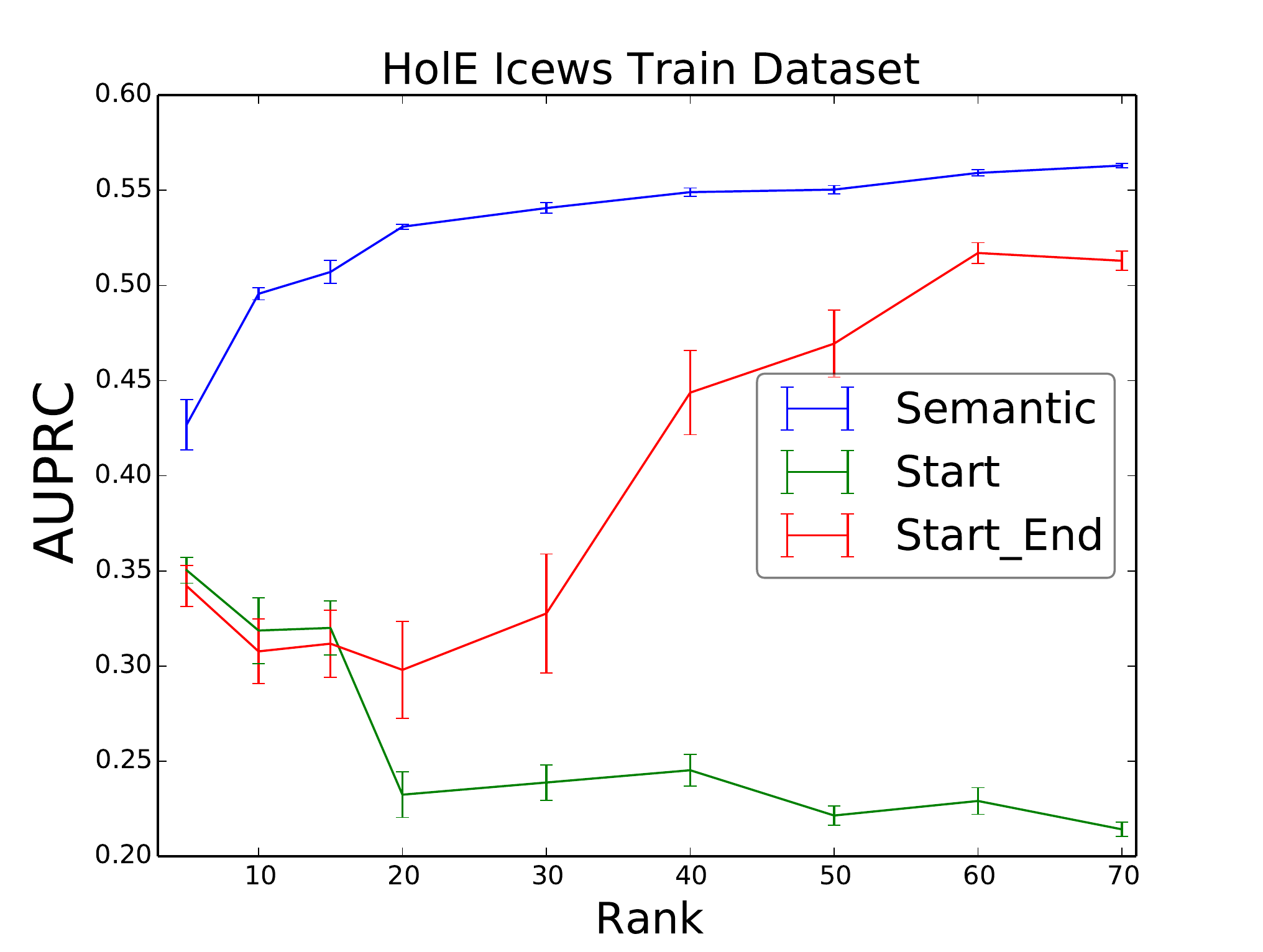}
  \caption{\small AUPRC scores vs. rank for the episodic-to-semantic projection on the ICEWS dataset with two different projection methods.}
  \label{fig:Projection_auprc}
\end{figure*}

To evaluate the projection, we compute the recall and area under precision-recall-curve (AUPRC) scores for the projection at different ranks on the ICEWS training dataset, and compare them with the scores obtained from training the semantic tensor separately. The semantic dataset contains positive triples, which are episodic events that continue until the last (current) timestamp, e.g. (Ant\'onio Guterres, SecretaryOf, UN, \emph{True}), along with negative triples extracted from already terminated episodic events, e.g. (Ban Ki-moon, SecretaryOf, UN, \emph{False}). During the test phase of projection, a triple from the semantic dataset is given with non-specified time index, e.g. $(e_s, e_p, e_o, \emph{True}/\emph{False}, t)$. Then, for the first method considering only the starting point of an episodic event, the projection to semantic space is computed as
\begin{equation}
  \theta^{proj}_{s,p,o} = [\sum_{t_{start}=1}^{T}\textbf{a}(e_{t_{start}})] \cdot \tilde{f},
  \label{eq:projection_start}
\end{equation}
while for the second method considering both starting and terminal points, the projection is computed as
\begin{equation}
  \theta^{proj}_{s,p,o} = \left[ \sum_{t_{start}=1}^T \textbf{a}(e_{t_{start}}) - \sum_{t_{end}=1}^T \textbf{a}(e_{t_{end}}) \right] \cdot \tilde{f}\ .
  \label{eq:projection_start_end}
\end{equation}
Then, the scores are evaluated by taking the label of the given semantic triple as the target, and taking $\theta^{proj}_{s,p,o}$ as the prediction. The goal of this test is to check how well the algorithms can project a given consecutive event $(e_s, e_p, e_o, t_{start} \cdots t_{end})$ to semantic knowledge space using only the marginalized latent representation of time. All other experimental settings are similar to those in Section~\ref{sec:Episodic_exp}, and the experiments were repeated four times on different sampled training datasets.

Figure~\ref{fig:Projection_recall} shows the recall scores for the two different projection methods on the training dataset in comparison to the separately trained semantic dataset. Due to limited space, we only show four models: ConT, Tucker, ComplEx, and HolE. As we can see, only the marginalization considering both starting and terminal time indices allows a reasonable projection from episodic memory to the \emph{current} semantic memory. Again, ConT\footnote{Note that since ConT doesn't have a direct semantic counterpart, we instead use the semantic results obtained using RESCAL. This is reasonable since ConT can be viewed as a high-dimensional (i.e., episodic) generalization of RESCAL.} exhibits the best performance, with its recall score saturating after $\tilde{r}\approx 15$. In contrast, HolE shows insufficient projection quality with sizable errors, especially at small ranks, which is due to its higher-order encoding noise. To show that the two types of latent representations of time do not simply eliminate each other for a correct episodic projection, Figure~\ref{fig:Projection_auprc} shows the AUPRC scores evaluated on the training dataset. Overall, this experiment supports the idea that semantic memory is a long-term storage for episodic memory, where the exact timing information is lost.

\begin{table*}[t]

  \setlength{\tabcolsep}{0.15pc}
  \footnotesize

  \caption{Filtered and raw Hits@10 scores for the episodic-to-semantic projection. Two projection methods, Start (Eq.~\ref{eq:projection_start}), Start-End (Eq.~\ref{eq:projection_start_end}), are compared. Furthermore, semantic ICEWS dataset with genuine semantic triples, and semantic ICEWS dataset with false triples are used for the projection experiments. Various projection scores are compared with the scores which are obtained by directly modeling the semantic ICEWS dataset with genuine semantic triples.}
  \label{table:projection_hits10}
  \centering
  \begin{tabular}{lcccccccccccc|cc}
    \toprule
      & \multicolumn{2}{c}{Start}  & &  \multicolumn{2}{c}{Start-End}  & &  \multicolumn{2}{c}{Start (false)}  & &  \multicolumn{2}{c}{Start-End (false)}  & &  \multicolumn{2}{c}{Semantic}  \\
    \cmidrule{2-3}  \cmidrule{5-6}  \cmidrule{8-9}  \cmidrule{11-12}  \cmidrule{14-15}
   Method & Filter & Raw  & &  Filter & Raw    & &  Filter & Raw   & &  Filter & Raw   & &  Filter & Raw   \\
   \midrule
  DistMult  &   3.8  & 3.6    & &  5.6  & 5.0    & &   4.0  & 3.8    & &   3.8  & 3.6    & &   59.3 & 32.4 \\
  HolE      &   5.8  & 5.4    & &  5.5  & 5.1    & &   4.7  & 4.5    & &   5.6  & 5.2    & &   56.1 & 31.3 \\
  ComplEx   &   4.1  & 3.7    & &  4.9  & 4.4    & &   3.9  & 3.7    & &   3.8  & 3.6    & &   60.1 & 29.4 \\
  Tucker    &   14.8 & 13.1   & &  15.1 & 13.4   & &   11.3 & 10.3   & &   11.8 & 10.9   & &   46.5 & 23.7 \\
  \midrule
  ConT      &   30.9 & 24.6   & &  \textbf{40.8} & \textbf{30.3}   & &   23.0 & 19.9   & &   22.6 & 19.3   & &   43.8 & 20.4 \\
  \bottomrule
  \end{tabular}
\end{table*}
\normalsize

For a fair comparison, in the last experiment we report the recall scores of the semantic models obtained by projecting the episodic models with respect to the temporal dimension. We compare two projection methods, the Start projection which only considers the staring point of episodic events (see Eq.~\ref{eq:projection_start}), and the Start-End projection which takes both the starting and terminal points of episodic events into consideration. In addition, we report the recall scores on two semantic datasets. The first one contains genuine semantic facts, while the second dataset contains false semantic triples which should already be ruled out through the projection.

Two different projections are performed on two semantic datasets, the genuine one and the false one. Theoretically, the recall scores on the genuine semantic dataset should be higher than those on the false dataset. Thus, the model hyper-parameters are chosen by monitoring the difference between the recall scores Hits@10 on the genuine and false semantic datasets.

Table.~\ref{table:projection_hits10} reports the filtered and raw Hits@10 metrics for different models, projection methods, and datasets. Moreover, we also compare the projection with the recall scores obtained by directly modeling the genuine semantic dataset using the corresponding semantic models~\footnote{Note that we use the RESCAL model as the corresponding semantic model for the ConT.}. The ConT model has the best projection performance, since its projected recall scores on the genuine dataset are much higher than those obtained on the false semantic dataset. Moreover, the Start-End projection method based on the ConT model is the only combination which achieves similar results compared to the corresponding semantic model.
One can also notice that all the projected compositional models are only able to tell whether a semantic triple is already ruled out or not before the last timestamp, however they can not provide good inference results on the genuine semantic dataset.

\section{Conclusion}
\label{sec:Conclusions}

This paper described the first mathematical models for the declarative memories: the semantic and episodic memory functions. To model these cognitive functions, we generalized leading approaches for static knowledge graphs (i.e., Tucker, RESCAL, HolE, ComplEx, DistMult) to 4-dimensional temporal/episodic knowledge graphs. In addition, we developed two novel generalizations of RESCAL to episodic tensors, i.e., Tree and ConT. In particular, ConT has superior performance overall, which indicates the importance of introduced high-dimensional latent representation of time for both sparse episodic tensor reconstruction and generalization.

Our hypothesis is that perception includes an active semantic decoding process, which relies on latent representations of entities and predicates, and that episodic and semantic memories depend on the same decoding process. We argue that temporal knowledge graph embeddings might be models for human cognitive episodic memory and that semantic memory (facts we know) can be generated from episodic memory by a marginalization operation. We also test this hypothesis on the ICEWS dataset, the experiments show that the \emph{current} semantic facts can only be derived from the episodic tensor by a proper projection considering both starting and terminal points of consecutive events.

\textbf{Acknowledgements.} This work is funded by the \textit{Cognitive Deep Learning} research project in Siemens AG.
\clearpage

\section*{References}

\bibliography{Reference}

\begin{thebibliography}{10}
\bibitem{singhal2012introducing}
A.~Singhal, Introducing the knowledge graph: things, not strings, Official
  google blog.
\end{thebibliography}

\end{document}